\documentclass{article} 
\usepackage{iclr2019_conference,times}

\usepackage{amsmath,amsfonts,bm}

\def\eqref#1{(\ref{#1})}

\def\1{\bm{1}}

\DeclareMathAlphabet{\mathsfit}{\encodingdefault}{\sfdefault}{m}{sl}
\SetMathAlphabet{\mathsfit}{bold}{\encodingdefault}{\sfdefault}{bx}{n}

\usepackage{booktabs}
\usepackage{hyperref}
\usepackage{url}
\usepackage[]{algorithm2e}
\usepackage{graphicx}
\usepackage{multirow}
\usepackage{subcaption}
\usepackage{todonotes}
\usepackage{mdwlist}
\usepackage{microtype}
\newcommand{\ignore}[1]{}
\newcommand{\nppmt}{NP$^2$MT}
\newcommand{\vect}[1]{\mathbf{#1}}

\title{\mbox{Neural Phrase-to-Phrase Machine Translation}}

\author{Jiangtao Feng\thanks{This work was done when Jiangtao and Jiayuan interned in Google.}\\ 
Fudan University \\
\texttt{fengjt16@fudan.edu.cn}
\And {Lingpeng Kong, Po-Sen Huang} \\
\quad \quad \quad \quad \ \  DeepMind \\
\texttt{\{lingpenk, posenhuang\}@google.com}
\And {Chong Wang, Da Huang} \\
\quad \quad \quad \quad Google \\
\texttt{\{chongw, dahua\}@google.com}
\And Jiayuan Mao$^*$ \\
Tsinghua University \\
\texttt{mjy14@mails.tsinghua.edu.cn}
\And Kan Qiao, Dengyong Zhou \\\quad \quad \quad \quad \  Google \\
\texttt{\{qiaokan, dennyzhou\}@google.com}
}

\iclrfinalcopy 
\begin{document}

\maketitle

\begin{abstract}
In this paper, we propose  Neural Phrase-to-Phrase Machine Translation (\nppmt). Our model uses a phrase attention mechanism to discover relevant input (source) segments that are used by a decoder to generate output (target) phrases. We also design an efficient dynamic programming algorithm to decode segments that allows the model to be trained faster than the existing neural phrase-based machine translation method by \cite{huang2018towards}. Furthermore, our method can naturally integrate with external phrase dictionaries during decoding. Empirical experiments show that our method achieves comparable performance with the state-of-the art methods on benchmark datasets. However, when the training and testing data are from different distributions or domains, our method performs better.
\end{abstract}

\section{Introduction}
Statistical phrase-based models used to be the state-of-the-art machine translation systems \citep{koehn2007moses}  before the deep learning revolution.  In contrast to word-based systems \citep{koehn2003statistical,lopez2008statistical,koehn2009statistical},  phrase-based approaches explicitly model phrase structures in both source and target sentences and their corresponding alignments. These linguistic structures can be understood as a form of inductive bias to the model, a key factor to its superior performance over word-based counterpart.

In recent years, we have witnessed the surge of neural sequence to sequence (seq2seq) models~\citep{bahdanau2014neural,sutskever2014sequence}. These models 
usually consist of three main components: an encoder that encodes the source sentence into a fixed-length vectoR, a decoder that generates the translation word by word, and an attention module that selectively retrieves the source side information as per the decoder's needs. Innovations in both architectures \citep{vaswani2017attention,gehring2017convolutional} and training techniques \citep{vaswani2017attention, ba2016layer} keep advancing the-state-the-art results on standard benchmarks for machine translation.

Despite the remarkable success of neural sequence to sequence models, the biases induced from phrase structures, which have been shown to be  useful in machine translation \citep{koehn2009statistical}, are largely ignored. Until recently, \cite{huang2018towards} developed Neural Phrase-based Machine Translation (NPMT) to incorporate target-side phrasal information into a neural translation model.
Their model builds upon the Sleep-WAke Networks (SWAN), a segmentation-based sequence modeling technique described in~\citet{wang2017sequence}. 

In this paper, we propose Neural Phrase-to-Phrase Machine Translation (\nppmt). 
Our contributions are twofold. First, we develop an explicit phrase-level attention mechanism to capture the source side phrasal information and their alignments with the target side phrase outputs. This approach also avoids the more costly dynamic programming procedure due to the monotonic requirement for input-output alignments in NPMT. {\nppmt} can achieve comparable performance with the state-of-the art methods on benchmark datasets.
Second, we can naturally incorporate an external phrase-to-phrase dictionary during the decoding procedure. NMT systems trained on a fixed amount of parallel data are known to have limitations with out-of-vocabulary words~\citep{luong2014addressing}. In other words, those out-of-vocabulary words are from different data distributions other than the training data. The standard remedy for this is to apply a post-processing step that patches in unknown words based on the word-level alignment information from attention mechanism~\citep{luong2014addressing,hashimoto2016domain}. In our model, given the phrase-level attentions, we develop a dictionary look-up decoding method with an external phrase-to-phrase dictionary. We demonstrate that our {\nppmt} model consistently outperforms Transformer~\citep{vaswani2017attention}
in a simulated open vocabulary setting and a cross domain translation setting.

\section{Neural phrase-to-phrase machine translation}

In this section, we first describe the proposed \nppmt{} model with the phrase-level attention mechanism. We show how it avoids the more costly dynamic programming used in NPMT \citep{huang2018towards}. Next, given the phrase-level attention mechanism, we develop a decoding algorithm that can naturally use an external phrase-to-phrase dictionary.



\subsection{\nppmt}
\begin{figure}
  \centering
    \includegraphics[width=1.0\textwidth]{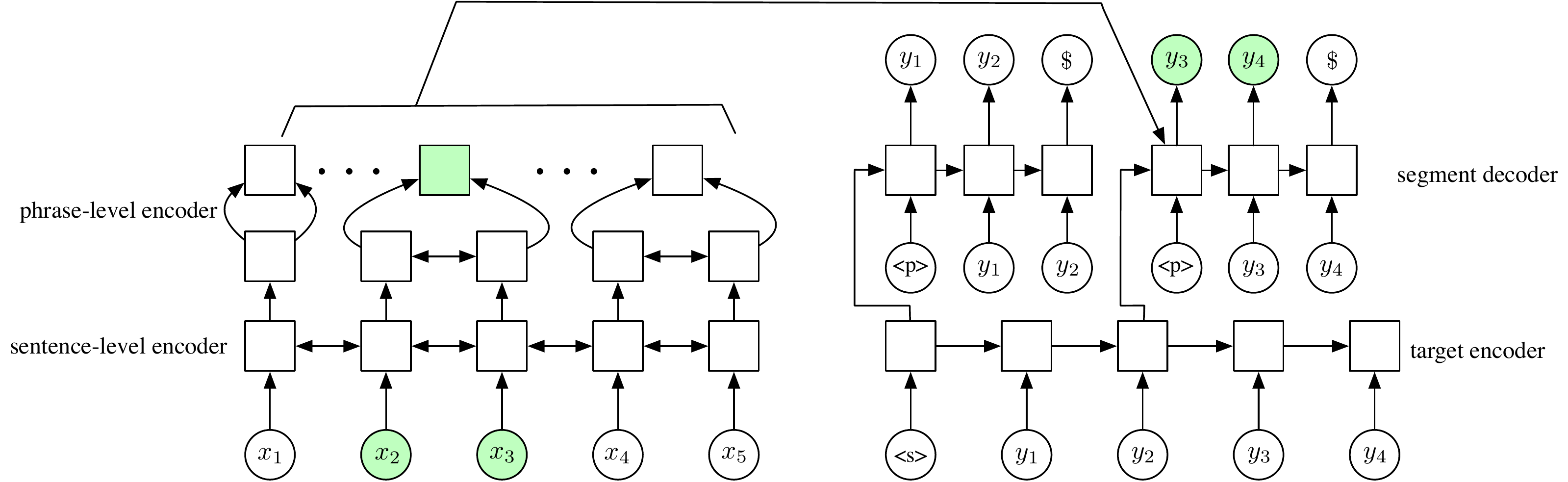}
\caption{The architecture of \nppmt{} model. The example shows how the target phrase $y_{3:4}$ is translated conditioning directly on the source phrase $x_{2:3}$, using the phrase-level attention. Note that for brevity, in the phrase-level encoder, we only show one possible segmentation of the source sentence. We use ``$\cdots$'' to indicate all the possible segments $x_{i:j}$ in Eq. \ref{eq:segs}.
}
  \label{fig:arch}
\end{figure}

Figure \ref{fig:arch} shows an overview of \nppmt{}.  There are three main components in our model:  a source encoder $f^{\text{src}}$, a target encoder  $f^{\text{tgt}}$, and a segment decoder $g$. The source encoder $f^{\text{src}}$ consists of a sentence-level and phrase-level encoders ($f_{\text{sent}}^{\text{src}}$, $f_{\text{phrase}}^{\text{src}}$), namely $f^\text{src}=f_{\text{phrase}}^{\text{src}} \circ f_{\text{sent}}^{\text{src}}$.  We use bidirectional LSTM~\citep{hochreiter1997long} for source encoders, LSTM for target encoder and the Transformer \citep{vaswani2017attention} for the segment decoder.\footnote{In our proposed model, we first tried LSTM to use as the targer decoder, but we later found Transformer did better to capture meaningful output segments.}

Consider a source sentence $x_{1:T}=\{x_1, x_2, \ldots, x_T\}$ and a target sentence $y_{1:T'}=\{y_1, y_2, \ldots, y_{T'}\}$. First, the source sentence $x_{1:T}$ is passed through sentence-level encoder $f_{\text{sent}}^{\text{src}}$ to obtain the word vector representation,
\begin{align}
\vect{s}_{1:T} = f_{\text{sent}}^{\text{src}}(x_{1:T}).
\end{align}
Second, we use phrase-level encoder $f^{\text{phrase}}_{\text{src}}$ to independently encode all possible segments $x_{i:j}$ in the source sentence as,
\begin{align}
    \vect{h}_{[i,j]} = f_{\text{phrase}}^{\text{src}}(\vect{s}_{i:j}), \forall (i,j) \in \{(i,j): 1 \leq i \leq j \leq i + L - 1 \leq T\}.
    \label{eq:segs}
\end{align}
where $L$ is the maximum segment length we will consider in our model. This architecture of computing phrase embedding is inspired by the Segmental RNNs \citep{kong2015segmental}, which has been shown to be effective in both text and speech tasks \citep{lu2017multitask}.

Denote $\vect{h}$ as the entire set of phrase-level encoding vectors. Now the rest of the model is more conveniently described as a generative process. Let $\pi(\cdot)$ be a concatenation operator, $\$$ be the end of segment symbol and $eos$ as the end of sentence symbol. 
\begin{enumerate*}
    \item For segment index $n=1,...,\infty$
    \begin{enumerate*}
        \item Update the attention state given all previous segments,
        \begin{align}
           \vect{a}_j = {\rm attn}(\vect{h}, \pi(z_{1:n-1})),\label{eq:attn} 
           \end{align}
           where $j= \sum_{m=1}^{n-1}|z_m|$ is the total length of all previous segments. Note that different collections of previous segments can lead to the same attention state as long as their concatenation is the same.
        \item Given the current state $\vect{a}_{j}$, sample words from segment decoder $g$  until we reach the end of segment symbol $\$$ or end of sentence $eos$. This gives us a segment $z_n$.
        \item If we see $eos$ in $z_n$, stop.
    \end{enumerate*}
    \item Concatenate $\{z_1,\ldots,z_N\}$ to obtain the output sentence $\pi(z_{1:N}) = y_{1:T'}$.
\end{enumerate*}
Since there are more than one way to obtain the same $y_{1:T'}$ using the generative process above, we need to consider all cases. Let $\vect{z}$ be a valid segmentation of $y_{1:T'}$, then we have
\begin{align}
    p(y_{1:T'}|x_{1:T}) =\sum_{\vect{z}: \pi(\vect{z}) = y_{1:T'}} \prod_{n=1}^{|\vect{z}|} p(z_n|\vect{a}_j),
\end{align}
where $|\vect{z}|$ is the number of segment in $\vect{z}$, attention $\vect{a}_j(z_{1:n-1}, \vect{h})$ is defined in Eq.~\ref{eq:attn} and the segment probability $p(z_n|\vect{a}_j)$ is defined via the decoder $g$. 

\subsection{Training}

Given a pair of source and target sentence $(x_{1:T}, y_{1:T'})$, the objective function of machine translation model is defined as follows,
\begin{equation}
\mathcal{L}(\vect{\theta})=-\log p(y_{1:T'}|x_{1:T}). \label{eq:loss}
\end{equation}Similar to NPMT in~\citet{huang2018towards}, direct computing Eq.~\eqref{eq:loss} is intractable. We also need to develop a dynamic programming algorithms to efficiently compute the loss function.
We denote conditional probability $p(y_{1:j}|x_{1:T})$ as $\alpha(j)$; then we have $p(y_{1:T'}|x_{1:T}) = \alpha(T')$. It can be shown that we have the following recursion, 
\begin{align*}
\alpha(j)& =\sum_{j'<j}{\alpha(j')p(y_{j'+1:j}|\vect{a}_{j'})},  \\
p(y_{j'+1:j}|\vect{a}_{j'}) & =p(\$|y_{j'+1:j},\vect{a}_{j'})p(y_{j'}|\vect{a}_{j'})\prod_{j'<k \le j}{p(y_k|y_{j':k-1},\vect{a}_{j'})},
\end{align*}
The initial condition is $\alpha(0)=1$. The computational complexity of this algorithm is to $O(T'L)$, where $L$ denotes the maximum segment length. Comparing with the original NPMT model \citep{huang2018towards} with $O(TT'L)$ training complexity, the proposed approach is much faster.

\subsection{Decoding}

For the proposed \nppmt{}, besides the standard beam search algorithm, we propose a method to decode with an external phrase-to-phrase dictionary. We leave the beam search algorithm in Appendix.

\paragraph{Dictionary integration in decoding.}

\begin{figure}
  \centering
    \includegraphics[width=0.65\textwidth]{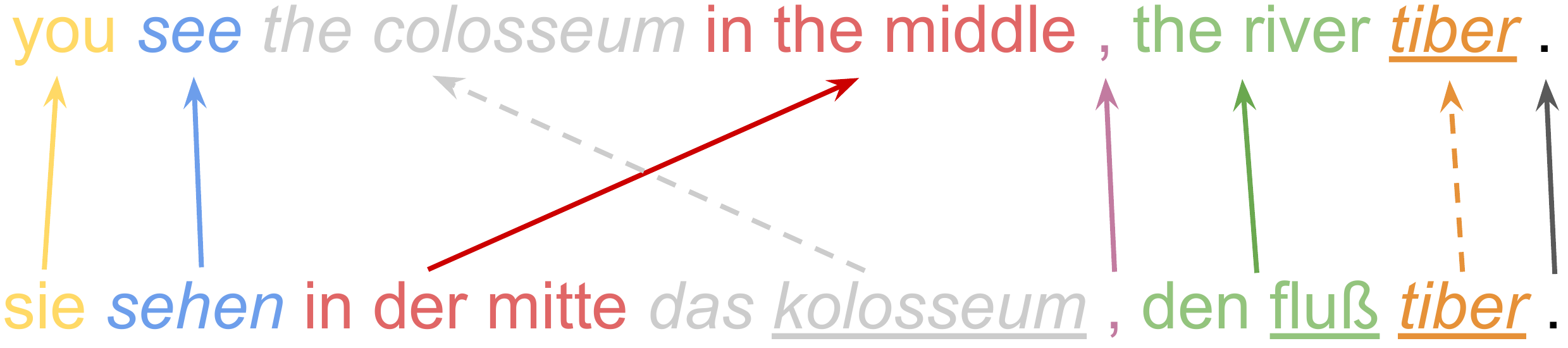}
\caption{The decoding process of \nppmt{} model with dictionary extension. The example shows how a German sentence ``sie sehen in der mitte das kolosseum , den fluß tiber .'' is translated into the English sentence ``you see the colosseum in the middle , the river tiber .'' Different colors are used to represent the aligned phrases. The underlined words (``kolosseum'', ``fluß'', ``tiber'') are OOV words, and the italic phrases (``sehen'', ``das kolosseum'', ``tiber'') can be found in the dictionary. The solid lines are used to denote the translation by the neural model, and the dashed ones are by the dictionary.}
  \label{fig:decode}
\end{figure}

The proposed \nppmt{} builds a phrase-to-phrase framework, which could be easily extended to leverage an external phrase-to-phrase dictionary. Here, we describe the dictionary-enhanced decoding algorithm in the greedy search case.\footnote{Integration dictionary to beam search is nontrivial. For the current model, the usage of the dictionary is determined by rules instead of probability scores. But beam search requires a quantitative comparison among the candidates consisting of the phrases generated by both of the rule-based system and the neural model. It is unclear how to quantify the former into beam search scores. We leave it for future work.}

The decoding process is shown by Figure \ref{fig:decode}. At each timestamp in the decoding process, the \nppmt{} will produce a phrase-level attention, and the phrase with the maximum attention is the one to be translated. Given the attended phrase, the model needs to decide whether to use a dictionary to translate the phrase or not. Here only the attended phrases with unknown words are translated with the dictionary if the dictionary includes its translation. The intuition is that when a person is reading a sentence, he/she decides to use a dictionary if he/she meets an unknown word and the attended phrase can be found in the dictionary. When there are multiple translations for a phrase in the dictionary, we use the model to score all candidates, and then choose the best one. Here in \nppmt{}, we use the segment decoder to do so. 
The decoding algorithm is shown in Algorithm \ref{alg:dict}

\begin{algorithm}[h]
 \KwData{a source sentence: $x_{1:T}$, a dictionary: $D$}
 \KwResult{a target sentence: $y$}
 Compute hidden representation $\vect{h}$ for the source sentence\;
 Set initial output segment $z^\text{tgt}=null$\;
 Set output sentence $y=\O$\;
  \While{eos $\notin$  $z^\text{tgt}$}{
   Compute attention state $\vect{a}_j$ and the attention distribution\;
   Select the attended source phrase $z^\text{src}$ with the maximum attention score\;
   \eIf{UNK $\in$ $z^\text{src}$ \&\& $z^\text{src}$ $\in$ $D$}{
       Retrieve all target translations $D(z^\text{src})$ from the dictionary\;
       Select the best candidate $z^\text{tgt}$ in $D(z^\text{src})$ by using the neural segment decoder $g$ as the scoring function
   }
   {
      Generate the next output segment $z^\text{tgt}$ by using the neural segment decoder $g(\vect{a}_j$)\;
   }
   Append $z^\text{tgt}$ to output $y$\;
  }
 \caption{\nppmt{} greedy decoding with dictionaries.}
\label{alg:dict}
\end{algorithm}


\section{Experiments}

In this section, we demonstrate \nppmt{} on several machine translation benchmarks under different settings. 
We first compare the \nppmt{} model with some strong baselines on IWSLT14 German-English, IWSLT 2015 English-Vietnamese machine translation tasks \citep{cettolo2014report, cettolo2015iwslt}. Furthermore, to demonstrate the benefit of using the phrase-level attention mechanism with an external dictionary, we evaluate \nppmt{} on open vocabulary and cross-domain translation tasks.\footnote{We focus on word-based models in this paper as words are more appropriate for dictionary representation.  We leave the BPE types of representations \citep{sennrich2015neural} in future work.}




\subsection{IWSLT Machine Translation Tasks}
\label{sec:iwslt}
In this section, we compare \nppmt{} with the state of the art machine translation models on IWSLT 14 German-English and IWSLT 15 English-Vietnamese machine translation tasks \citep{cettolo2014report,cettolo2015iwslt}.
The IWSLT14 German-English machine translation data \citep{cettolo2014report} comes from translated TED talks, and the dataset
contains roughly 153K training sentences, 7K development sentences, and 7K test sentences. We use the same preprocessing and dataset splits as in \citep{ott2018scaling}. Sentences longer than $175$ words are removed from the dataset.
For the IWSLT 15 English-Vietnamese machine translation task, the data is from translated TED talks, and the dataset contains roughly 133K training sentence pairs provided by the IWSLT 2015 Evaluation Campaign \citep{cettolo2015iwslt}. 
Following the same preprocessing steps in \citet{luong17, huang2018towards}, we use the TED tst2012 (1553 sentences) as the validation set for hyperparameter tuning and TED tst2013 (1268 sentences) as the test set.
We report the results on the test set.





We use 6-layer BiLSTMs to encode at both words and segment level in the source encoder, 6-layer LSTMs as the target encoder, and a 6-layer transformer as the segment decoder.\footnote{We also experiment on LSTM segment decoder variant, but the performance is inferior.}
The models are trained using Adam~\citep{kingma2014adam}. 
Similar to the RNMT+ learning rate scheduler \citep{chen2018best, ott2018scaling}, we use a three-stage learning rate scheduler by replacing the exponential decay with a linear one for fast convergence. Specifically, the learning rate is quickly warmed up to the maximum, kept at the maximum value to $50\%$ progress, and finally decaying to zero. In our experiments, we set the maximum learning rate to $1e^{-3}$, weight decay to $1e^{-4}$, the word embedding dimensionality to $256$, and the dropout rate to $0.4$. 
The performances of baseline models, including BSO \citep{wiseman2016sequence}, Seq2Seq with attention, Actor-Critic \citep{bahdanau2016actor}, \citet{luong_mt_2015} and NPMT \citep{huang2018towards} are taken from \citep{huang2018towards}.
We use the fairseq implementation of Transformer\footnote{\href{https://github.com/pytorch/fairseq}{https://github.com/pytorch/fairseq}} and set the number of layers in both encoder and decoder to 6.\footnote{No further improvement is observed by increasing the depth.} We fine-tuned the hyperparameters using the valid set and found the best-performed hyperparameters, except the maximum learning rate, are the same as in \nppmt{}. We replace the default inverse-square-root scheduler \citep{vaswani2017attention} with the proposed learning rate scheduler which works better empirically and the maximum learning rate of Transformer is set to $2e^{-4}$.

The IWSLT 14 German-English and IWLST 15 English-Vietnamese test results are shown in Tables \ref{table:iwslt_de-en} and \ref{table:iwslt_en-vi}.
The proposed \nppmt{} achieves comparable results as the transformer \citep{vaswani2017attention}, and outperforms other baseline models in both tasks.
Besides, comparing with the NPMT model, \nppmt{} achieves better results with a much less training time (24 hours vs 2 hours on a Nvidia V100 GPU).

\begin{table}[h]
 \centering
 \begin{tabular}[h]{lcc}
 \toprule
 \multirow{2}{*}{Model} & \multicolumn{2}{c}{BLEU} \\
 & Greedy & Beam Search \\
 \midrule
 BSO \citep{wiseman2016sequence} & 23.83 & 25.48 \\
 Seq2Seq with attention & 26.17 & 27.61 \\
 Actor-Critic \citep{bahdanau2016actor} &27.49 &28.53 \\
 Transformer \citep{vaswani2017attention} & 31.27 & 32.30 \\
 \midrule
 NPMT \citep{huang2018towards} & 28.57 & 29.92\\
 \midrule
 \nppmt{} & 30.99 & 31.70 \\
 \bottomrule
 \end{tabular}
 \caption{Performance on IWSLT14 German-English test set. }
  \label{table:iwslt_de-en}
\end{table}

\begin{table}[h!]
	\centering
	\begin{tabular}{lcc}
		\toprule		
		\multirow{2}{*}{Model} & \multicolumn{2}{c}{BLEU } \\ 
		&  Greedy & Beam Search \\ 	
		\midrule

		\citet{luong_mt_2015} & - &  23.30 \\    
		Seq2Seq with attention & 25.50 &  26.10 \\  
        Transformer \citep{vaswani2017attention} & {29.72} & 30.74 \\	\midrule
        NPMT \citep{huang2018towards} & {26.91} & {27.69} \\	\midrule
        \nppmt{} & 29.93 & 30.60 \\
		\bottomrule
	\end{tabular}
  \caption{Performance on IWSLT15 English-Vietnamese test set. 
  }
	\label{table:iwslt_en-vi}
\end{table}

\subsection{Translation with Dictionary}
To demonstrate the usefulness of using an external dictionary, we set up an experiment to show how an external bilingual dictionary can enhance the performance of \nppmt{} in the open vocabulary translation tasks, where there are a large number of out-of-vocabulary (OOV) words. 

We build the open vocabulary machine translation task by using UNK to replace the infrequent words. By infrequent words, we mean the words that appear fewer times than a predefined threshold in the training data.
For the masked words, the model can only see their corresponding texts, where no vector representations are available. 
Given this setting, the models are supposed to translate the sentences with masked unknown words.

In our experiment, we evaluate different thresholds of infrequency words (3, 5, 10, 20, 50, 100), to show how using an external dictionary can improve translation in the open vocabulary setting. In Tables \ref{table:iwslt-de-en-oov} and \ref{table:iwslt-en-vi-oov}, we report the results on IWSLT14 German-English and IWSLT15 English-Vietnamese datasets.
For both datasets, we use Moses \citep{koehn2007moses} to extract an in-domain dictionary $D_{I}$ from the training data, including $97,399$ and $84,817$ translations.
Besides, we also try an out-of-domain dictionary on IWSTL14 German-English task, which is extracted from the WMT14 German-English dataset, and includes $1,243,722$ translations.
The goals of these two tasks are different. For IWSLT15 English-Vietnamese task, the dictionary is extracted from the training data by Moses, and is an in-domain one. The external resource here is the dictionary extract tool, namely Moses. But for IWSLT14 German-English taks, the dictionary itself is an external resource which can be obtained from any textual corpus in any way, and is an out-of-domain one. In other words, the difference between these task lies in the overlapping between the training data and the dictionary. 
Note that in these dictionaries, there could be several translations attached to each word. For comparison, we also report the transformer with a trivial lookup mechanism by translating the attended unknown word with the given dictionary.

\begin{table}[h]
 \centering
  \begin{tabular}[h]{c|c|c|c|c|c|c|c}
  \toprule
  \multicolumn{2}{c|}{Threshold} & 3 & 5 & 10 & 20 & 50 & 100 \\
  \midrule
  \multirow{2}{*}{Vocab Size} & German & 35,479 & 23,327 & 13,679 & 8,055 & 3,831 & 2,152 \\
   & English & 24,743 & 17,807 & 11,559 & 7,359 & 3,807 & 2,224 \\
  \hline
  \multirow{2}{*}{Test Data OOV Rate} & German & 3.8\% & 4.7\% & 6.2\% & 8.1\% & 11.6\% & 14.8\%\\
  & English & 1.5\% & 2.1\% & 3.0\% & 4.4\% & 7.2\% & 10.2\% \\
  \midrule 
  \multicolumn{2}{c|}{Transformer} & 31.27 & 30.92 & 30.35 & 28.88 & 26.27 & 23.73 \\
  \multicolumn{2}{c|}{Transformer + $D_{I}$} & 31.27 & 30.97 & 30.65 & 29.34 & 27.78 & 26.01 \\
  \multicolumn{2}{c|}{Transformer + $D_{O}$} & \textbf{31.67} & 31.40 & 31.04 & 29.43 & 27.42 & 25.20 \\
  \hline
  \multicolumn{2}{c|}{\nppmt{}} & 30.99 & 30.92 & 29.86 & 28.29 & 25.81 & 23.08 \\
  \multicolumn{2}{c|}{\nppmt{} + $D_{I}$} & 30.99 & 31.01 & 30.33 & 29.52 & \textbf{28.02} & \textbf{26.61} \\
  \multicolumn{2}{c|}{\nppmt{} + $D_{O}$} & 31.48 & \textbf{31.75} & \textbf{31.06} & \textbf{30.04} & 27.86 & 25.85 \\
  \bottomrule
  \end{tabular}
  \caption{IWSLT14 German-English Translation with in-domain and out-of-domain dictionaries.}
  \label{table:iwslt-de-en-oov}
\end{table}


\begin{table}[h]
 \centering
  \begin{tabular}[h]{c|c|c|c|c|c|c|c}
  \toprule
  \multicolumn{2}{c|}{Threshold} & 3 & 5 & 10 & 20 & 50 & 100 \\
  \midrule
  \multirow{2}{*}{Vocab Size} & English & 24,415 & 17,191 & 10,919 & 6,799 & 3,431 & 1,943 \\
   & Vietnamese & 10,663 & 7,711 & 5,343 & 3,863 & 2,575 & 1,943 \\
  \hline
  \multirow{2}{*}{Test Data OOV Rate} & English & 2.2\% & 2.8\% & 3.9\% & 5.6\% & 8.9\% & 12.1\% \\
  & Vietnamese & 0.8\% & 1.0\% & 1.4\% & 2.0\% & 3.3\% & 4.6\% \\
  \midrule
  \multicolumn{2}{c|}{Transformer} & 29.72 & 29.71 & 28.97 & 28.40 & 25.68 & 23.00 \\
  \multicolumn{2}{c|}{Transformer + $D_{I}$} & 29.72 & 29.72 & 29.01 & \textbf{28.71} & 26.92 & 25.02 \\
  \hline
  \multicolumn{2}{c|}{\nppmt{}} & 29.93 & 29.78 & 29.27 & 27.87 & 25.89 & 23.63 \\
  \multicolumn{2}{c|}{\nppmt{}+$D_{I}$} & \textbf{29.94} & \textbf{29.79} & \textbf{29.36} & 28.38 & \textbf{27.65} & \textbf{26.31} \\
  \bottomrule
  \end{tabular}
  \caption{IWSLT15 English-Vietnamese Translation with an in-domain dictionary.}
  \label{table:iwslt-en-vi-oov}
\end{table}

\begin{figure}[h]
\centering
\begin{subfigure}{.5\textwidth}
  \centering
\includegraphics[width=1\linewidth]{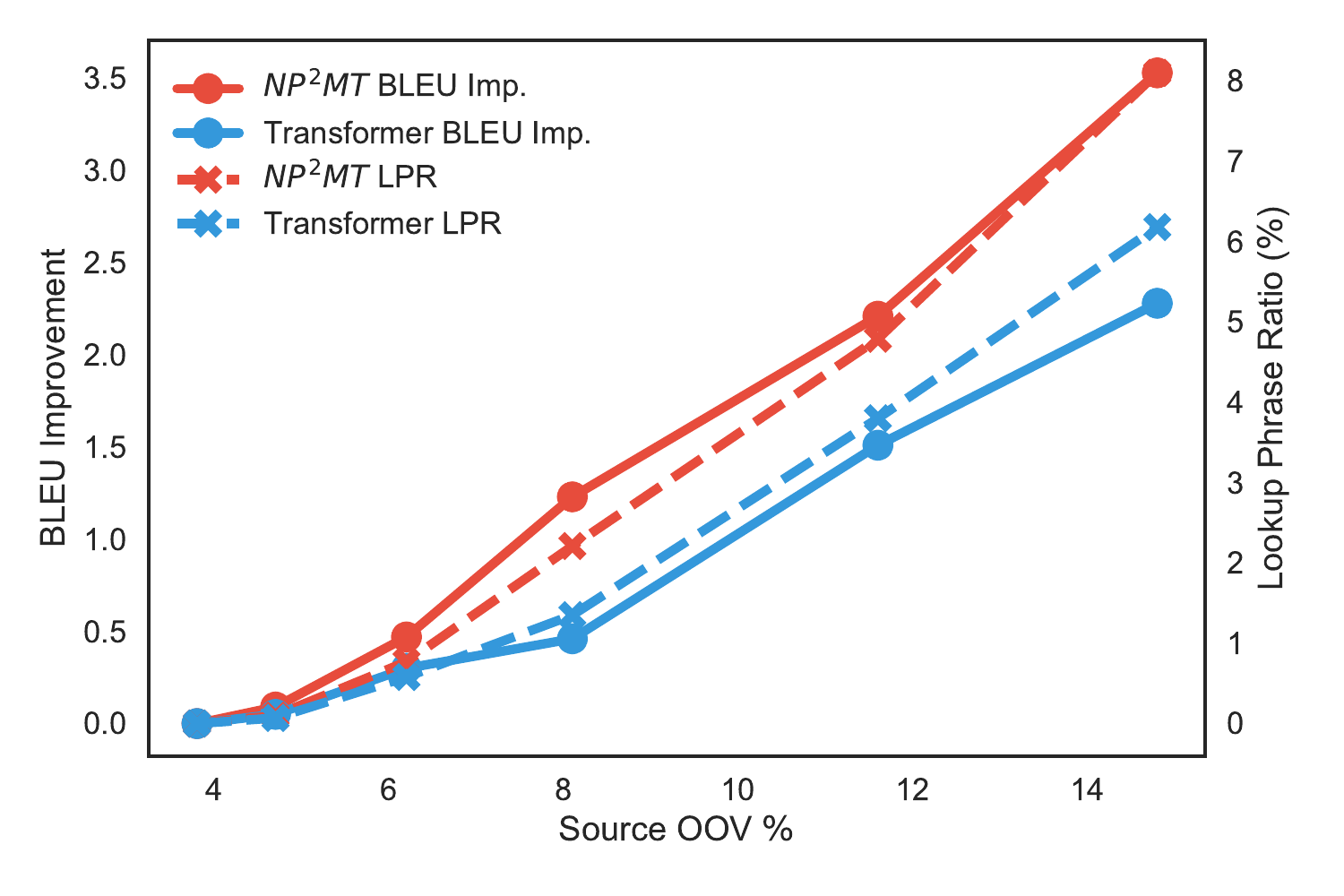}
  \caption{IWSLT14 DE-EN}  
\end{subfigure}%
\begin{subfigure}{.5\textwidth}
  \centering
  \includegraphics[width=1\linewidth]{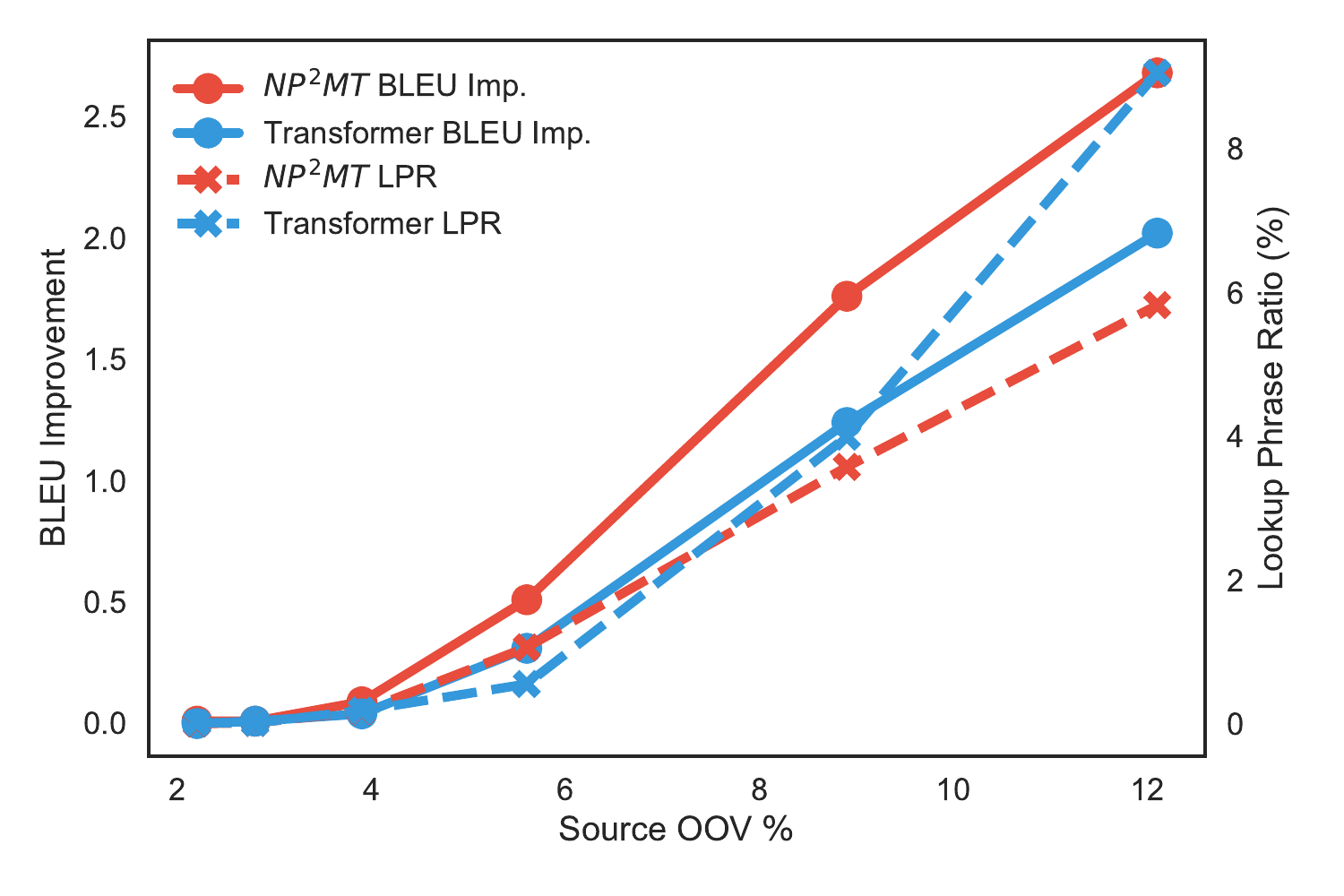}
  \caption{IWSLT15 EN-VI}
\end{subfigure}
\caption{Comparison on BLEU score improvement and lookup phrase ratio under different source language OOV rates both under in-domain dictionaries. }
\label{fig:phrase_table_lookup_ratio}
\end{figure}
\begin{figure}[h]
\centering
\begin{subfigure}{.5\textwidth}
  \centering
\includegraphics[width=1\linewidth]{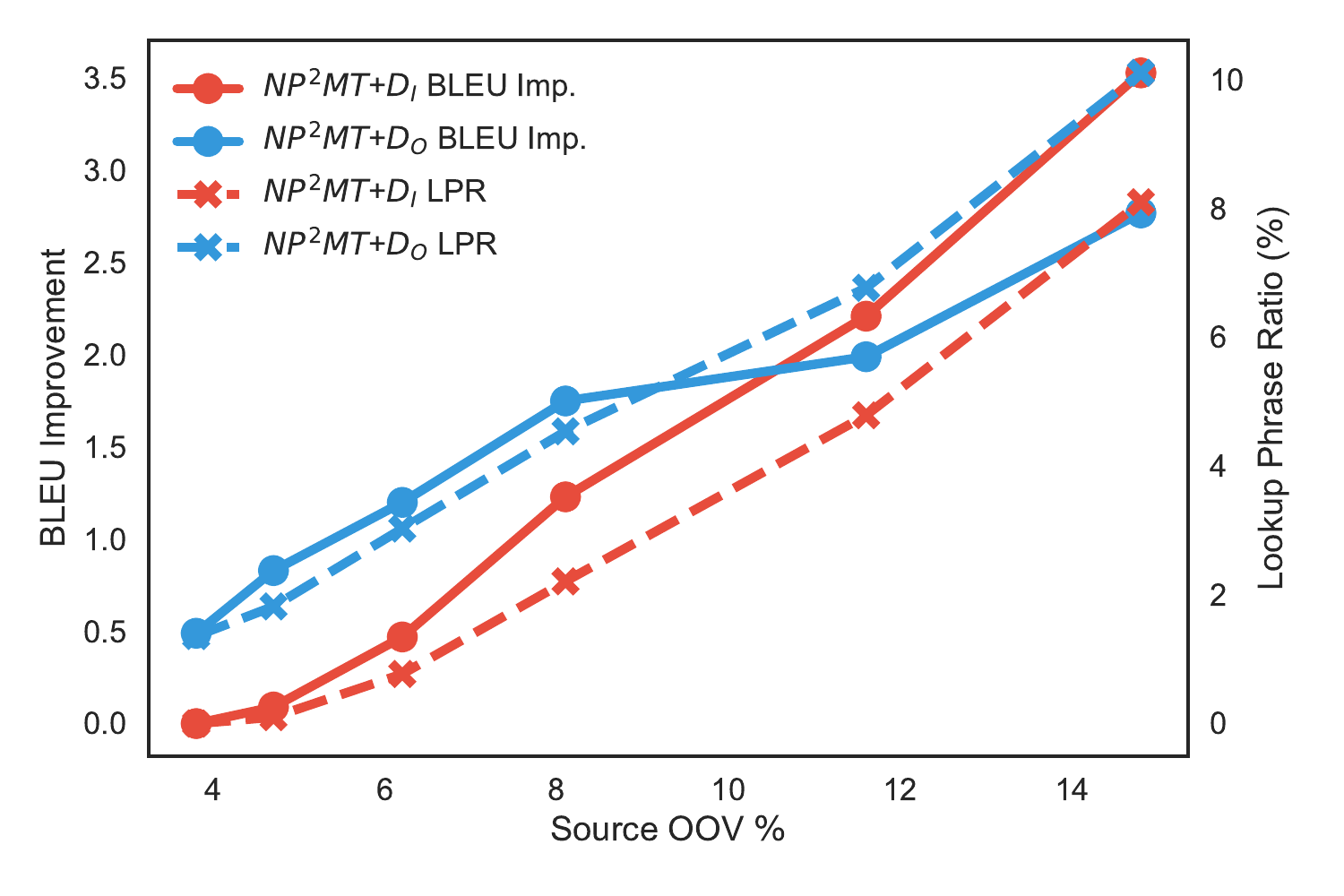}
\end{subfigure}%
\caption{Comparison on BLEU score improvement and lookup phrase ratio  for IWSLT14 DE-EN under in-domain and out-of-domain dictionaries. }
\label{fig:wmt-iwslt-phrase-table}
\end{figure}

In Fig. \ref{fig:phrase_table_lookup_ratio} and Fig. \ref{fig:wmt-iwslt-phrase-table}, we analyze the lookup phrase ratio and the BLEU score improvement on both datasets, where the lookup phrase ratio is the percentage of target words that is translated using the dictionary.  We observe the lookup phrase ratio and BLEU improvement are increasing when the source OOV rate is higher. This also suggests the need of using the dictionary in a high OOV scenario. 
%
In Fig. \ref{fig:phrase_table_lookup_ratio}, when there are $12-14\%$ OOV rate, using an in-domain dictionary can enhance the performance of the \nppmt{} model by $3.5$ and $2.6$ BLEU scores on IWSLT14 German-English and IWSLT15 English-Vietnamese task, whereas only $2.3$ and $2.0$ BLEU score improvement is achieved with the transformer model on these task respectively. 
Fig. \ref{fig:wmt-iwslt-phrase-table} shows the difference between a small in-domain dictionary $D_I$ and a large out-of-domain one $D_O$ for IWSLT14 DE-EN. When the OOV rate is low, more improvements are gained by larger $D_O$ because it contained more translations. But when the OOV rate becomes higher, $D_I$ show its effectiveness by providing accurate in-domain translation candidates.

Table \ref{table:de-en-example} shows the translation examples by \nppmt{} with unknown word frequency threshold $50$.
At the word level, \nppmt{} discovers some meaningful phrases like ``this is'', ``the machine'' and ``read by''. For unseen words, such as ``genf'', ``bescheidene'' and ``zeitschriften'', could be attended and decoded with the dictionary. 

\begin{table}[h]
\centering
\begin{tabular}[htp]{l|l}
\toprule
\multirow{1}{*}{Source} & $ \text{[dies]}_1 \; \text{ist} \; \text{[die maschine]}_2 \; \text{[unterhalb]}_3 \; \text{(von \underline{genf})}_4 \; \text{[.]}_5 $ \\
Greedy decoding & $\text{[this is]}_1 \; \text{[the machine]}_2 \; \text{[below]}_3 \; \text{(geneva)}_4 \; \text{[.]}_5$ \\
\multirow{1}{*}{Target ground truth}  & this is the machine below \underline{geneva} . \\
\midrule
Source & $\text{[dies]}_1 \; \text{ist} \; \text{(eine \underline{bescheidene})}_2 \; \text{[kleine]}_3 \; \text{(\underline{app})}_4 \; \text{[.]}_5 $ \\
Greedy decoding & $\text{[this is]}_1 \; \text{(a modest)}_2 \; \text{[little]}_3 \; \text{(app)}_4 \; \text{[.]}_5$ \\
Target ground truth  & this is a \underline{modest} little app . \\
\midrule
Source & $ \text{[unsere]}_1 \; \text{(\underline{zeitschriften})}_2 \; \text{[werden von]}_3 \; \text{[millionen]}_4 \; \text{[gelesen]}_5 \; \text{[.]}_6 $ \\
Greedy decoding & $\text{[our]}_1 \; \text{(journals)}_2 \; \text{[are]}_3 \; \text{[read by]}_5 \; \text{[millions]}_4 \; \text{[.]}_6 $ \\
Target ground truth  & our \underline{magazines} are read by millions . \\
\bottomrule
\end{tabular}
\caption{DE-EN translation examples, where ``[$\cdot$]'' represents the phrase boundary, ``($\cdot$)'' represents the phrase is looked up by the external dictionary, and ``\underline{~~$\cdot$~~}'' represents the frequency of the word is under the vocabulary threshold and hence the word is replaced by the ``UNK'' token in the network vocabulary. The subscript represents the corresponding phrases found by \nppmt{}. }
\label{table:de-en-example}
\end{table}




\subsection{Cross-Domain Translation}

In this section, Transformer and the proposed model are evaluated on a cross-domain machine translation task. 
The models are trained and validated on the IWSLT14 German-English dataset, which is originally from TED talks, and tested on the WMT14 German-English dataset, which is extracted from the news domain.

In this task, we used the lowest threshold $3$ for the training data to make the observed vocabulary as complete as possible. Note that although the used vocabulary is the most complete one for training domain, the OOV rate is still high in the out-of-domain test data. And the dictionary, which is extracted from the test domain, is used to handle such OOV words at the word/phrase level.\footnote{In practice, it would be better to use expert-created dictionaries for cross-domain tasks.}

In Table \ref{table:iwslt-wmt-de-en}, we show the experimental results with and without dictionary using Transformer and \nppmt{} model.
We observed that while dictionary contributed to the performance of both models, the ``\nppmt{} + $D_{\text{WMT}}$'' model used the dictionary more often and achieved higher BLEU gain (+$1.25$) than transformer did (+$0.91$) in the experiments. 



\begin{table}[h]
 \centering
  \begin{tabular}[h]{c|c}
  \toprule
  Model & BLEU \\
  \midrule
  Transformer & 14.69 \\
  Transformer + $D_{\text{WMT}}$ & 15.60 \\
  \hline
  \nppmt{} & 14.86 \\
  \nppmt{}+$D_{\text{WMT}}$ & \textbf{16.11} \\
  \bottomrule
  \end{tabular}
  \caption{Results on cross-domain translation by training the model on IWSLT14 and testing on WMT14, both with German-English data. In the test data from WMT14, OOV rates are $12.8$\% for German and $6.7$\% for English.}
  \label{table:iwslt-wmt-de-en}
\end{table}

\section{Related Work}
Neural phrase-based machine translation is first introduced by \citet{huang2018towards}. The model builds upon Sleep-WAke Networks (SWAN), a segmentation-based
sequence modeling technique described in \citep{wang2017sequence} and mitigate
this issue of monotonic alignment assumption by introducing a new layer to perform (soft) local reordering on input sequences. Another related model shares the monotonic alignment assumption (which is often inappropriate in many language pairs) is 
the segment-to-segment neural transduction model (SSNT) \citep{yu2016online,yu2016neural}. Our model relies on the phrasal attention mechanism rather than marginalize out the monotonic alignments using dynamic programming.
There have been several works that propose different ways to incorporate phrases into attention based 
neural machine translation. \citet{Wang2017emnlp}, \citet{Tang2016neural} and \citet{zhao2018phrase} incorporate the phrase table as memory in the neural machine translation architecture. \citet{hasler2018neural} use a user-provided phrase table of terminologies into NMT system by organizing the beam search into multiple stacks corresponding to subsets of satisfied constraints as defined by FSA states. \citet{dahlmann-EtAl-EMNLP2017} divides the beams into the word beam and the phrase beam of fixed size. \citet{he2016improved} uses statistical machine translation (SMT) as features in the NMT model under the log-linear framework. \citet{yang2018modeling} enhance the self-attention networks to capture useful phrase patterns by imposing learned Gaussian biases.
\citet{nguyen2018phrase} also incorporates phrase-level attention with transformer by encoding a fixed number of n-grams (e.g. unigram, bigram). 
However, \citet{nguyen2018phrase} only focuses on phrase-level attention on the source side, whereas our model focuses on phrase-to-phrase translation by attending on a phrase at the source side and generating a phrase at the target side. 
Our model further integrates an external dictionary during decoding which is important in open vocabulary and cross domain translation settings. 

\section{Conclusions}
We proposed Neural Phrase-to-Phrase Machine Translation (\nppmt{}) that uses a phrase-level attention mechanism to enable  phrase-to-phrase level translation in a neural machine translation system. By using this phrase-level attention, we can incorporate an external dictionary during decoding.  We show that we can  improve the machine translation results in open vocabulary and cross domain settings by using the external phrase dictionary in the decoding time.

\section*{Acknowledgements}
We thank Dani Yogatama and Chris Dyer for helpful comments and dicussions on this project.

\bibliography{iclr2019_conference}
\bibliographystyle{iclr2019_conference}

 \appendix
 
\section{Beam Search}

The beam search algorithm is mostly based on \citep{wang2017sequence}. Here we use a word-level beam search algorithm shown in Algorithm \ref{alg:beam}.

\begin{algorithm}[h]
 \KwData{a source sentence: $x_{1:T}$, beam size $k$}
 \KwResult{a target sentence: $y$}
 compute the hidden representation $\vect{h}$ for the source sentence\;
 prev\_sent $\gets$ end-of-sentence token eos\;
 prev\_segment $\gets$ end-of-segment token $\$$\;
 init\_score  $\gets$ $1$\;
 // Z is the appendable candidates\;
 // Y is the finished candidates\;
 $Z=\{(\text{prev\_sent}, \text{prev\_segment}, \text{init\_score})\}$\;
 $Y=\O$\;
 \While{Z does not reach maximum length}{
   // $\tilde{Z}$ is the next appendable candidates\;
   $\tilde{Z}=\O$\;
   \For {\text{prev\_sent}, \text{prev\_segment}, \text{score} in $Z$}{
    Generate $k$ next\_tokens with probability $p_k$\;
    \For {\text{next\_token} in \text{next\_tokens candidate set}} {
        \If{\text{next\_token} is eos} {
            prev\_segment $\gets$ prev\_segment + next\_token\;
            prev\_sent $\gets$ prev\_sent + prev\_segment\;
            score $\gets$ score + $p_k$\;
            Append (prev\_sent, score) to $Y$\;
        }
        \If{\text{next\_token} is $\$$} {
            prev\_sent $\gets$ prev\_sent + prev\_segment\;
            score $\gets$ score + $p_k$\;
            Append (prev\_sent, $\$$, score) to $\tilde{Z}$\;
        }
        \If {\text{next\_token} is not $\$$ or eos}{
            prev\_segment $\gets$ prev\_segment + next\_token\;
            prev\_sent $\gets$ prev\_sent + prev\_segment\;
            score $\gets$ score + $p_k$\;
            Append (prev\_sent, prev\_segment, score) to $\tilde{Z}$\;
        }
    }
   }
   Keep only top $k$ candidates in $\tilde{Z}$\;
   $Z \gets \tilde{Z}$
 }
 $y$ $\gets$ the best candidates with highest score in $Y$
  \caption{\nppmt{} Beam Search.} \label{alg:beam}
\end{algorithm}

\end{document}